%% file: acl.tex
\pdfoutput=1

\documentclass[11pt]{article}

\usepackage{acl}

\usepackage{times}
\usepackage{latexsym}

\usepackage[T1]{fontenc}

\usepackage[utf8]{inputenc}

\usepackage{microtype}

\usepackage{graphicx}
\usepackage{booktabs}
\usepackage{tabularx,paralist}
\usepackage{amsmath}

\newcommand{\norm}[1]{\left\|#1\right\|}

\usepackage{prerex}
\usepackage{verbatim}
\usepackage{tikz-dependency}
\usepackage{amssymb}
\usetikzlibrary{arrows,automata}
\usetikzlibrary{positioning}

\title{On the Sensitivity and Stability of Model Interpretations in NLP}

\author{Fan Yin, Zhouxing Shi, Cho-Jui Hsieh, and Kai-Wei Chang \\
  University of California, Los Angeles \\
  \texttt{\{fanyin20, zshi, chohsieh, kwchang\}@cs.ucla.edu};
  }
\date{}

\begin{document}
\maketitle
\begin{abstract}
Recent years have witnessed the emergence of a variety of post-hoc interpretations that aim to uncover how natural language processing (NLP) models make predictions. Despite the surge of new interpretation methods, it remains an open problem how to define and quantitatively measure the \textit{faithfulness} of interpretations, i.e., to what extent interpretations reflect the reasoning process by a model. We propose two new criteria, sensitivity and stability, that provide complementary notions of faithfulness to the existed removal-based criteria. Our results show that the conclusion for how faithful interpretations are could vary substantially based on different notions. Motivated by the desiderata of sensitivity and stability, we introduce a new class of interpretation methods that adopt techniques from adversarial robustness. Empirical results show that our proposed methods are effective under the new criteria and overcome limitations of gradient-based methods on removal-based criteria. Besides text classification, we also apply interpretation methods and metrics to dependency parsing. Our results shed light on understanding the diverse set of interpretations. 
\end{abstract}

\section{Introduction}
As complex NLP models are widely deployed in real-world applications, there is an increasing interest in understanding how these models come to certain decisions. As a result, the line of research on interpretation techniques grows rapidly, facilitating a broad range of model analysis, from building user trust on models \citep{Ribeiro2016Why, hase2020evaluating} to exposing subtle biases \citep{Zhao2017men, Doshi2017Towards}.

In this paper, we focus on \textit{post-hoc interpretations} in NLP. Given a trained model and a specific input text, post-hoc interpretations assign an importance score to each token in the input which indicates its contribution to the model output. Current methods in this direction can be roughly divided into three categories: gradient-based methods \citep{Simonyan13saliency, Li16Visual}; reference-based methods \citep{Sundararajan17Axio, Shrikumar17Learning}; and perturbation-based methods \citep{ZeilerF14, Ribeiro2016Why}.

Despite the emergence of new techniques, one critical issue is that there is little consensus on how to define and evaluate the faithfulness of these techniques, i.e., whether they reflect the true reasoning process by a model. A widely employed criterion, especially in NLP, is the \textit{removal-based} criterion \citep{ deyoung2020eraser}, which removes or only preserves a set of tokens given by interpretations and measures how much the model prediction would change. However, as pointed out in prior work \citep{bastings2020elephant, AnconaCO018}, the corrupted version of an input produced during evaluations falls out of the distribution that models are trained on, and thus results in an inaccurate measurement of faithfulness. This limitation prevents removal-based metrics from being used as the golden standard for evaluating interpretations. To remedy this, we complement the removal-based criterion with two other criteria, \textit{sensitivity} and \textit{stability}, which are overlooked in prior works. 


\textit{Sensitivity} is based on the notion that models should be more sensitive to perturbations on tokens identified by a faithful explanation. In contrast to the removal-based criterion, which completely removes important tokens, the sensitivity criterion adds small but adversarial perturbations in a local region of the token embedding, and thus preserves the structure of input sentences as well as interactions between context words. This criterion is recently discussed in \citet{Hsieh2021Evaluations} in computer vision, while we provide comprehensive analyses on various NLP models and tasks. Note that while the removal-based criterion asks the question: \textit{if some important tokens did not `exist', what would happen}, the sensitivity criterion asks: \textit{if some important tokens were `changed' adversarially, what would happen}.

\textit{Stability} assumes that a faithful interpretation should not produce substantially different explanations for two inputs that the model finds similar. There are several attempts to generate such a pair of inputs. The most relevant one is \citet{GhorbaniAZ19}. However, their method is only applicable to differentiable interpretations. Our work proposes a new paradigm based on adversarial word substitution that employs a black-box algorithm to generate a semantically related neighbor of the original input, which is specially designed for NLP and applicable to all interpretations techniques.

The above two metrics highlight the connection between interpretability and robustness. Experiments show that interpretations which perform well on the removal-based criterion might not do well on the new criteria. Motivated by the limitations of existing interpretations and the desiderata of sensitivity, we propose \emph{robustness-based methods}, based on projected gradient descent (PGD) attacks \citep{Madry18Towards} and certifying robustness~\citep{jia2019certified,huang2019achieving,Shi2020Robust, Xu2020Auto}. We demonstrate that the new methods achieve top performance under sensitivity and stability. Moreover, as a simple improvement to gradient-based methods, our methods avoid the gradient saturation issues of gradient-based methods under the removal-based criterion.

Another limitation of removal-based metrics emerges when interpreting dependency parsing -- when input tokens are removed, the tree structure is drastically changed and a model might not be able to produce a meaningful parse tree. Thus, there are little discussion for dependency parsing interpretations. In this paper, we propose a new paradigm to interpret dependency parsers leveraging prepositional phrase (PP) attachment ambiguity examples. To our best knowledge, this is the first work to study interpretations on dependency parsing. We demonstrate that sensitivity does not change the output tree structure as much as removal-based ones do, and provide analyses for interpretation methods with our paradigm and metrics.




Our contributions can be summarized as follows.

\begin{compactenum}
    \item We discuss two overlooked notions of faithfulness in NLP interpretations. Our notions emphasize the connection between interpretability and robustness. We systematically evaluate interpretations under these notions, including existed removal-based ones. The code for this paper could be found at \url{https://github.com/uclanlp/NLP-Interpretation-Faithfulness}.
    \item We propose new robustness-based interpretations inspired by the sensitivity metric and demonstrate their effectiveness under both sensitivity and stability.
    \item We propose a novel paradigm to evaluate interpretations on the dependency parsing task.
\end{compactenum}

\section{Faithfulness Evaluation Criteria}
\label{EvalMethod}
A faithful post-hoc interpretation identifies the important parts of the input a model prediction relies on. Let $x = \left[x_1; x_2;  \ldots; x_n\right]$ be a sequence of tokens. $e\left(\cdot \right)$ denotes the token embedding function. An NLP model $f$ takes the embedding matrix $e\left( x \right) \in \mathcal{R}^{n \times d}$ as input and provides its prediction $f\left(e\left( x \right)\right)=y$. Let $s_y\left(e\left(x\right)\right)$ denote the output score of $f\left(e\left( x \right)\right)$ on $y$. The exact form of $s_y\left(e\left(x\right)\right)$ is defined in Appendix \ref{Appendix4: task details}. An interpretation assigns an importance score to each token which indicates its contribution to the model decision.

We first review the well-established \emph{removal-based criterion} and emphasize its relation to the two criteria defined in this paper 1) \emph{sensitivity}, and 2) \emph{stability}, for which we propose novel paradigms to adapt them to various NLP tasks.

\noindent{\bf Removal-based Criterion} A well-established notion of interpretation faithfulness is that the presence of important tokens should have more meaningful influence on the model's decision than random tokens, quantified by the removal-based criterion. We adopt the \textit{comprehensiveness} and the \textit{sufficiency} score in \citet{deyoung2020eraser}. The comprehensiveness score measures how much the model performance would drop after the set of ``relevant" tokens identified by an interpretation is removed. A higher comprehensiveness score suggests the tokens are more influential to the model output, and thus a more faithful explanation. The sufficiency score measures to what extent the original model performance is maintained when we solely preserve relevant tokens. A lower sufficiency score means less change in the model prediction, and thus a more faithful explanation. See \citet{deyoung2020eraser} for detailed definitions. Note that completely removing input tokens produces incomplete texts. Large perturbation of this kind lead to several issues as pointed out by prior studies \citep{feng2018pathologies, bastings2020elephant}.




\noindent{\bf Ours: Sensitivity} Instead of removing important tokens, the sensitivity criterion adds \emph{local} but adversarial noise to embedding vectors of the important tokens and measures the magnitude of the noise needed to change the model prediction. This is inspired by the notion that models should be more sensitive to perturbations being added to relevant tokens compared to random or irrelevant tokens. From the adversarial robustness perspective \citep{Hsieh2021Evaluations}, this notion implies that by perturbing the most relevant tokens, we can reach the local decision boundary of a model with the minimum perturbation magnitude.

Given the sequence of relevant tokens $r_k$, sensitivity adds perturbation to its embedding $e\left(r_k\right)$ but keeps the remaining token embeddings unchanged. Then, it measures the minimal perturbation norm, denoted as $\epsilon_{r_k}$, that changes the model prediction for this instance:
\begin{equation*}
    \epsilon_{r_k} = \mathop{\min}\norm {\boldsymbol{\delta_{r_k}}}_F \indent \text{s.t.} \indent f\left(e\left(x\right) + \boldsymbol{\delta_{r_k}} \right) \ne y,
\end{equation*}
where $\norm{\cdot}_F$ is the Frobenius norm of a matrix, and $\boldsymbol{\delta_{r_k}} \in \mathcal{R}^{n \times d}$ denotes the perturbation matrix where only the columns for tokens in $r_k$ have non-zero elements. Since the exact computation of $\epsilon_{r_k}$ is intractable, we use the PGD attack \citep{Madry18Towards} with a binary search to approximate $\epsilon_{r_k}$. A lower $\epsilon_{r_k}$ suggests a more faithful interpretation. In practice, we vary the size of $r_k$, compute multiple $\epsilon_{r_k}$, and summarize them with the area under the curve (AUC) score.

\paragraph{Ours: Stability} Another desired property of faithfulness is that a faithful interpretation should not give substantially different importance orders for two input points that the model finds similar. To construct a pair of similar inputs, we propose to generate contrast examples to the original one by synonym substitutions. A contrast example of $x$, $\tilde x$, satisfies (1) has at most $k$ different but synonymous tokens with $x$; (2) the prediction score at $\tilde x$ changes less than $\tau$ compared to the score at $x$. The goal of these two conditions is to generate (almost) natural examples where the changes of model outputs are smaller than a threshold $\tau$. Given all contrast examples, we search for the one that leads to the largest rank difference $\mathcal{D}$ between the importance order for $x$, $m\left(x\right)$ and the alternated order $ m\left(\tilde x\right)$:
\begin{equation*}
\begin{gathered}
        \mathop{\arg\max}\nolimits_{\tilde x} \mathcal{D}\left(m\left(x\right), m\left(\tilde x\right)\right), \\
    \text{s.t.} \, 
    \left|s_y\left(e\left(x\right)\right) - s_y\left(e\left(\tilde x\right)\right)\right| \leq \tau, 
    \,\norm{x - \tilde x}_0 \leq k.
\end{gathered}
\end{equation*}
Specifically, we first extract synonyms for each token $x_i$ following \citet{alzantot2018generating}. Then, in the decreasing order of $m\left(x\right)$, we greedily search for a substitution of each token that induces the largest change in $m\left(x\right)$ and repeat this process until the model output score changes by more than $\tau$ or the pre-defined constraint $k$ is reached. Finally, we measure the difference $\mathcal{D}$ between two importance ranks using Spearman’s rank order correlation \citep{spearman1961proof}. We call this criterion \textit{stability}. A higher score indicates that the ranks between this input pair are more similar, and thus a more faithful interpretation.

Note that instead of using the gradient information of interpretation methods to perturb importance ranks like \citet{GhorbaniAZ19}, our algorithm treats interpretations as black-boxes, which makes it applicable to non-differentiable ones. Also, compared to \citet{ding2021evaluating}, who manually construct similar input pairs, our method is a fully automatic one as suggested by their paper.



\section{Interpretations via Adversarial Robustness Techniques}
\label{intermethods}
Experiments indicate that existing methods do not work well with the sensitivity and stability metrics (Sec. \ref{TextClsResandDiscussion}). In this section, we define a new class of interpretation methods by adopting techniques in adversarial robustness to remedy this. We first give a brief review of existing interpretation approaches and then introduce our new methods.

\subsection{Existing Interpretation Methods}
We roughly divide the existing methods into three categories: \emph{gradient-based methods}, \emph{reference-based methods}, and \emph{perturbation-based methods}, and discuss the representatives of them. 

\noindent{\bf Gradient-based methods} The first class of methods leverage the gradient at each input token. To aggregate the gradient vector at each token into a single importance score, we consider two methods: 1) using the $L_2$ norm, $\norm {\frac{\partial s_y\left(e\left(x\right)\right)}{\partial e\left(x_i\right)}}_2$, referred to as \textbf{Vanilla Gradient} (VaGrad) \citep{Simonyan13saliency}, and 2) using the dot product of gradient and input, $\left(\frac{\partial s_y\left( e\left(x\right)\right)}{\partial e\left(x_i\right)}\right)^\top \cdot e\left(x_i\right)$, referred to as \textbf{Gradient $\cdot$ Input} (GradInp) \citep{ Li16Visual}.


\noindent{\bf Reference-based methods} These methods distribute the difference between model outputs on a reference point and on the input as the importance score for each token. We consider \textbf{Integrated Gradient} (IngGrad) \citep{Sundararajan17Axio} and \textbf{DeepLIFT} \citep{Shrikumar17Learning}. IngGrad computes the linear intergral of the gradients from the reference point to the input. DeepLIFT decomposes the difference between each neuron activation and its `reference activation' and back-propagates it to each input token. We use DeepLIFT with the Rescale rule. Note DeepLIFT diverges from IngGrad when multiplicative interactions among tokens exist ~\citep{AnconaCO018}.


\noindent{\bf Perturbation-based methods} Methods in this class query model outputs on perturbed inputs. We choose \textbf{Occlusion} \citep{ZeilerF14} and \textbf{LIME} \citep{Ribeiro2016Why}. Occlusion replaces one token at a time by a reference value and uses the corresponding drop on model performance to represent the importance of each token. LIME uses a linear model to fit model outputs on the neighborhood of input $x$ and represents token importance by the weights in the trained linear model. 

\subsection{Proposed Robustness-based Methods}
We propose two methods inspired from the PGD attack \citep{Madry18Towards} and the certifying robustness algorithms \citep{Xu2020Auto} in adversarial robustness. 

\noindent{\bf VaPGD and PGDInp} The PGD attack in adversarial robustness considers a small vicinity of the input and takes several ``mini-steps" within the vicinity to search for an adversarial example. Consider the token embeddings for the input $x$, we perform $t$ iterations of the standard PGD procedure starting from
$e^{\left(0\right)} = e\left(x\right)$:
\begin{equation*}
    e^{\left(j\right)} \!=\! \mathcal{P}\left(e^{\left(j \!-\! 1\right)} \!-\! \alpha \nabla s_y\left(e^{\left(j \!-\! 1\right)}\right)\right), \, j\!=\!1,2,\ldots, t.
\end{equation*}
$\mathcal{P}$ represents the operation that projects the new instance at each step back to the vicinity of $e\left(x\right)$, and $\alpha$ is the step size. 

Intuitively, $e^{\left(t\right)} - e\left(x\right)$ tells us the descent direction of model confidence. Similar to the gradient-based methods, the importance of each token $x_i$ can be either represented by $\norm{e^{\left(t\right)}_{i} - e\left(x_{i}\right)}_{2}$, where $e^{\left(t\right)}_{i}$ is the i-th column in $e^{\left(t\right)}$, referred to as \textbf{Vanilla PGD} (VaPGD), or by $\left(e\left(x_i\right)- e^{\left(t\right)}_{i}\right)^\top\cdot e\left(x_i\right)$, referred to as \textbf{PGD $\cdot$ Input} (PGDInp)

Note that different from the PGD attack we use for approximating the sensitivity criterion, we manually decide the magnitude of the vicinity of $e\left( x \right)$ instead of using a binary search. We add perturbations to the whole sentence at the same time. Also, the final $e^{\left(t\right)}$ does not necessarily change the model prediction. See Appendix \ref{Appendix2: interpdetails} for details.


\noindent{\bf Certify} Certifying robustness algorithms also consider a vicinity of the original input and aim to provide guaranteed lower and upper bounds of a model output within that region. We use the linear relaxation based perturbation analysis (LiRPA) discussed in \citep{Shi2020Robust, Xu2020Auto}. LiRPA looks for two linear functions that bound the model. Specifically, LiRPA computes
$\overline{\boldsymbol{W}}$, $\underline{\boldsymbol{W}}$,
$\overline{\boldsymbol{b}}$, and
$\underline{\boldsymbol{b}}$ that satisfy $\sum_i\underline{\boldsymbol{W_i}}e\left(x_i'\right) +\underline{\boldsymbol{b}} \leq s_y\left(e\left(x'\right)\right) \leq \sum_i \overline{\boldsymbol{W_i}}e\left(x'_i\right) +\overline{\boldsymbol{b}}$ for any point $e\left(x'\right)$ that lies within the $L_2$ ball of $e\left(x\right)$ with size $\delta$. We use the IBP+backward method in \citet{Xu2020Auto}. It uses Interval Bound Propagation \citep{gowal2018effectiveness,mirman2018differentiable} to compute bounds of internal neurons of the model and then constructs the two linear functions with a bound back-propagation process~\citep{zhang2018efficient,singh2019abstract}. Finally, the importance score of the $i$-th token in the input is represented by $\underline{\boldsymbol{W_i}} \cdot e\left(x_i\right)$, where $\underline{\boldsymbol{W_i}}$ is the $i$-th row of $\underline{\boldsymbol{W}}$. We call this method \textbf{Certify}.

\noindent{\bf Robustness-based vs. Gradient-based} Gradient-based methods provide a linear approximation of the model decision boundary at the single input, which is not accurate for non-linear models. Robustness-based methods instead search multiple steps in neighbors and approximate the steepest descent direction better. We also empirically show that robustness-based methods avoid the saturation issue of gradient-based methods, i.e, gradient becomes zero at some inputs. See Appendix \ref{Appendix8 Case study Saturation}. Note that VaPGD (PGDInp) degrades to VaGrad (GradInp) when the number of iterations is 1.

\noindent{\bf Robustness-based vs. IngGrad}
IngGrad leverages the average gradient in a segment between the input and a reference. It is likely to neglect local properties desired by the sensitivity criterion. Robustness-based methods instead search in the vicinity of the input, and thus local properties are better preserved. See results in Sec. \ref{TextClsResandDiscussion}.

\section{Experiments on Text Classification}
In this section, we present the results on text classification tasks under the three criteria. We find that the correlation between interpretation faithfulness based on different criteria are relatively low in some cases. Results verify the effectiveness of our new methods.

\subsection{Experimental Setup}
\label{ExpSetup}
\noindent{\bf Datasets} We conduct experiments on three text classification datasets: SST-2 \citep{socher2013recursive}, Yelp \citep{zhang2015character}, and AGNews \citep{zhang2015character} following \citet{jain2019attention}'s preprocessing approach. All of them are converted to binary classification tasks. SST-2 and Yelp are sentiment classification tasks where models predict whether a review is \textit{negative} (0) or \textit{positive} (1). AGNews is to discriminate between \textit{world} (0) and \textit{business} (1) articles. See Appendix \ref{Appendix1: datastatistics} for statistics of the three datasets. When evaluating interpretation methods, for each dataset, we select 200 random samples (100 samples from class 0 and 100 samples from class 1) from the test set.

\noindent{\bf Models} For text classification, we consider two model architectures: BERT \citep{devlin2019bert} and BiLSTM \citep{hochreiter1997long}.

\noindent{\bf Interpretation Methods} Besides our robustness-based interpretations \textbf{PGDInp}, \textbf{VaPGD}, and \textbf{Certify}, we experiment with six others from three existing categories: \textbf{VaGrad}, \textbf{GradInp} (gradient-based); \textbf{IngGrad}, \textbf{DeepLIFT} (reference-based); and \textbf{Occlusion}, \textbf{LIME} (perturbation-based). We also include a random baseline \textbf{Random} that randomly assigns importance scores. We use comprehensiveness (\textbf{Comp.}), sufficiency (\textbf{Suff.}), sensitivity (\textbf{Sens.}), and stability (\textbf{Stab.}) as metrics. 


See Appendix \ref{Appendix1: datastatistics}$\sim$\ref{Appendix3: evaldetails} for experimental details.

\subsection{Results and Discussion}
\label{TextClsResandDiscussion}

\begin{table*}[!t]
\renewcommand{\arraystretch}{0.8}
\centering
\small
\begin{tabular}{m{1.08cm}m{.75cm}<{\centering}m{.75cm}<{\centering}m{.75cm}<{\centering}m{.85cm}<{\centering}m{.75cm}<{\centering}m{.75cm}<{\centering}m{.75cm}<{\centering}m{.85cm}<{\centering}m{.75cm}<{\centering}m{.75cm}<{\centering}m{.75cm}<{\centering}m{.85cm}<{\centering}}
  \toprule
  & \multicolumn{4}{c}{\bf{SST-2}} & \multicolumn{4}{c}{\bf{Yelp}} & \multicolumn{4}{c}{\bf{AGNews}}
\\
\cmidrule{2-5}\cmidrule{6-9}\cmidrule{10-13}
  \bf{Methods} & \bf{Comp.$\uparrow$}  & \bf{Suff.$\downarrow$} & \bf{Sens.$\downarrow$} & \bf{Stab.$\uparrow$} & \bf{Comp.}  & \bf{Suff.} & \bf{Sens.} & \bf{Stab.} & \bf{Comp.}  & \bf{Suff.} & \bf{Sens.} & \bf{Stab.}\\
  \toprule
  Random & 0.202 & 0.412 & 0.853 & -0.343 &0.166 & 0.383 & 1.641 & -0.254 & 0.039 & 0.269 &1.790 & -0.392 \cr
  \midrule
  \rowcolor{lightgray}
  VaGrad & 0.371 & 0.286 & 0.546 & 0.850 & 0.273 & 0.254 & 1.034 & 0.798 & 0.251 & 0.113 &1.041 & 0.843 \cr
  \rowcolor{lightgray}
  GradInp & 0.257 & 0.371 &0.814 & 0.336 & 0.240 & 0.328 & 1.363 & 0.559 & 0.081 & 0.281 & 1.379 & 0.390 \cr
  \midrule
  Occlusion & 0.498 & 0.208 & 0.655 & 0.604 &0.480 &\textbf{0.192} & 1.135 & 0.662 & 0.233 & 0.169 & 1.330 & 0.609 \cr
  LIME & \textbf{0.562} & \textbf{0.208} & 0.626 & 0.458 & \textbf{0.511} & 0.199 & 1.260 & 0.002 & \textbf{0.461} & \textbf{0.063} &1.178 & 0.115 \cr
  \midrule
  \rowcolor{lightgray}
  IngGrad & 0.420 & 0.286 & 0.711 & 0.729 & 0.417 & 0.201 & 1.350 & 0.793 & 0.284 & 0.153 & 1.251 & 0.761 \cr
  \rowcolor{lightgray}
  DeepLIFT & 0.266 & 0.367 & 0.820 & 0.351 & 0.265 & 0.315  & 1.413 & 0.569 & 0.082 & 0.135 & 1.326 & 0.457 \cr
  \midrule
  PGDInp & 0.390 & 0.284 & 0.560 & 0.605 &0.275 &0.295 & 1.079 & 0.628 & 0.205 & 0.141 & 1.028 & 0.590 \cr
  VaPGD & 0.373 & 0.277 &\textbf{0.542} &\textbf{0.853} &0.285 &0.266 &\textbf{1.022} &\textbf{0.832} & 0.256 & 0.109 &\textbf{0.995} & \textbf{0.869}\cr
  \bottomrule\hline

\end{tabular}
\caption{Results of evaluating interpretations for BERT under three criteria on text classification datasets. $\uparrow$ means a higher number under this metric indicates a better performance. $\downarrow$ means the opposite. The best performance across all interpretations is \textbf{bolded}. 
\textit{Certify} is missed here since current certifying robustness approaches cannot be scaled to deep Transformer-based models like BERT. See statistical analyses on Appendix \ref{statisticaltesting}.}
\label{BERTTextClsRes}
\end{table*}

\begin{table*}[h!]
\renewcommand{\arraystretch}{0.8}
\centering
\small
\begin{tabular}{m{1.08cm}m{.75cm}<{\centering}m{.85cm}<{\centering}m{.55cm}<{\centering}m{.85cm}<{\centering}m{.75cm}<{\centering}m{.85cm}<{\centering}m{.65cm}<{\centering}m{.85cm}<{\centering}m{.75cm}<{\centering}m{.85cm}<{\centering}m{.65cm}<{\centering}m{.85cm}<{\centering}}
  \toprule
  & \multicolumn{4}{c}{\bf{SST-2}} & \multicolumn{4}{c}{\bf{Yelp}} & \multicolumn{4}{c}{\bf{AGNews}}
\\
\cmidrule{2-5}\cmidrule{6-9}\cmidrule{10-13}
  \bf{Methods} & \bf{Comp.$\uparrow$}  & \bf{Suff.$\downarrow$} & \bf{Sens.$\downarrow$} & \bf{Stab.$\uparrow$} & \bf{Comp.}  & \bf{Suff.} & \bf{Sens.} & \bf{Stab.} & \bf{Comp.}  & \bf{Suff.} & \bf{Sens.} & \bf{Stab.}\\
  \toprule
  Random & 0.162 & 0.291 & 5.394 & -0.316 & 0.035 & 0.217 & 14.242 & -0.242 & 0.062 &0.170 & 13.712 & -0.378 \cr
  \midrule
  \rowcolor{lightgray}
  VaGrad &0.196 & 0.256 & 3.448 &0.860 &0.139 &0.108 & 9.438 & 0.887 & 0.061 & 0.187 &  10.485 & 0.812 \cr
  \rowcolor{lightgray}
  GradInp & 0.520 & 0.036 & 4.327 & 0.692 & 0.610 & -0.057 & 11.719 & 0.810 & 0.345 & 0.006 & 13.286 & 0.773 \cr
  \midrule
  Occlusion & 0.595 & -0.006 & 4.436 & 0.756 & 0.750 &\textbf{-0.062} & 11.725 & 0.816 & 0.513 & -0.018& 12.573 & 0.753 \cr
  LIME & \textbf{0.609} & -0.001 & 4.367 & 0.563 & 0.378 & 0.013 & 12.504 & 0.137 & 0.591 & -0.021 & 11.915 & 0.292 \cr
  \midrule
  \rowcolor{lightgray}
  IngGrad & 0.606 & \textbf{-0.007} & 4.500 & 0.767 &\textbf{0.780} & -0.062 & 12.394 & 0.849 & \textbf{0.657} & \textbf{-0.021} & 12.608  & 0.815 \cr
  \rowcolor{lightgray}
  DeepLIFT & 0.538 & 0.024 & 4.404 & 0.669 & 0.637 & -0.059 & 11.738 & 0.816 & 0.381 & -0.014 & 12.146 & 0.735 \cr
  \midrule
  PGDInp & 0.548 & 0.008 & 4.228 & 0.713 & 0.663 & -0.058 & 11.247 & 0.806 & 0.430 & -0.006 & 11.302 & 0.794 \cr
  VaPGD & 0.229 & 0.214 & \textbf{3.420} &\textbf{0.875} & 0.166 & 0.094 &\textbf{8.943} & \textbf{0.901} & 0.113 & 0.113 & \textbf{9.740} & \textbf{0.815} \cr
  Certify & 0.524 & 0.038 & 4.317 & 0.692 & 0.612 & -0.056 & 11.738 & 0.811 & 0.367 & -0.011 & 12.143 &0.778 \cr
  \bottomrule\hline

\end{tabular}
\caption{Results of evaluating different interpretation methods for BiLSTM. Same symbols as above.}
\label{LSTMClsRes}
\end{table*}


\noindent{\bf Overall Results} Results of interpretations for BERT and BiLSTM are presented in Table \ref{BERTTextClsRes} and \ref{LSTMClsRes}. The interpretations' performance are averaged over three runs on models trained from different random seeds. Results verify the effectiveness of our proposed robustness-based methods. Specifically, VaPGD achieves the best performance under the sensitivity and the stability criteria for both BERT and BiLSTM. Our methods also outperform their gradient-based counterparts under removal-based criteria. Especially, when interpreting BERT on SST-2 and AGNews, GradInp has near random performance. PGDInp can avoid these unreasonable behaviors. See Appendix \ref{Appendix8 Case study Saturation} for a qualitative study on this, where we find PGDInp does not suffer from the saturation issue as GradInp. Also notice that the superior performance of robsutness-based methods are consistent on BERT and BiLSTM+GloVe, which demonstrate that it is not influenced by the embeddings being used.

However, the performance of other methods tend to be inconsistent under different measurements. For example, under the removal-based criterion, IngGrad performs well for BiLSTM, which gives four out of six best numbers. But, IngGrad has very limited performance under the sensitivity metric, especially for BiLSTM on SST-2 and Yelp. Similar issues exist for LIME and Occlusion. Also, one might fail to recognize the faithfulness of VaPGD by solely looking at the removal-based criterion. Thus, when deploying interpretation methods on real tasks, we advocate for a careful selection of the method you use based on the underlying faithfulness notion that aligned with your goal.



\colorlet{red1}{red}
\colorlet{red2}{red!50}
\colorlet{red3}{red!25}
\colorlet{blue1}{blue}
\colorlet{blue2}{blue!50}
\colorlet{blue3}{blue!25}
\makeatletter
\pgfdeclaregenericanchor{top base}{%
  \csname pgf@anchor@#1@north\endcsname
  \pgf@anchor@generic@top@base@main
}
\pgfdeclaregenericanchor{top base west}{%
  \csname pgf@anchor@#1@north west\endcsname
  \pgf@anchor@generic@top@base@main
}
\pgfdeclaregenericanchor{top base east}{%
  \csname pgf@anchor@#1@north east\endcsname
  \pgf@anchor@generic@top@base@main
}
\def\pgf@anchor@generic@top@base@main{%
  {%
    \pgfmathsetlength\pgf@ya{\pgfkeysvalueof{/pgf/outer ysep}}%
    \advance\pgf@y-\pgf@ya
    \pgfmathsetlength\pgf@ya{\pgfkeysvalueof{/pgf/inner ysep}}%
    \advance\pgf@y-\pgf@ya
    \pgf@ya=0pt
    \pgfutil@loop
    \ifdim\pgf@y>\baselineskip
      \advance\pgf@y-\baselineskip
      \advance\pgf@ya\baselineskip
    \pgfutil@repeat
    \global\pgf@y=\pgf@ya
  }%
}
\makeatother
\newcommand\Textbox[2]{%
    \parbox[c][#1][c]{6.7cm}{#2}}
\begin{figure}[!t]

\tikzset{
    >=stealth',
    punkt/.style={
           very thick,
           rectangle split,
           rectangle split parts=2,
           inner ysep=1.2mm,
           inner xsep=1.5mm,
           rounded corners,
           draw=black, thick,
           text width=7.2cm},
    punx/.style={
           rectangle,
           rounded corners,
           align=left,
           text width=0.55cm,
           text depth=0.01cm,
           draw=white},
  header/.style = {%
    inner ysep = +1.5em,
    append after command = {
      \pgfextra{\let\TikZlastnode\tikzlastnode}
      node [header node] (header-\TikZlastnode) at (\TikZlastnode.north) {#1}
      node [span = (\TikZlastnode)(header-\TikZlastnode)]
        at (fit bounding box) (h-\TikZlastnode) {}
    }
  },
}

\begin{tikzpicture}[scale=1.0]

\node[punx](blank) {
\Textbox{0.4cm}{}
};
\node[punkt, below=0cm of blank](vapgd2){
\Textbox{0.3cm}{\small \colorbox{black}{\color{white} IngGrad} \quad \textbf{Comp.$\uparrow$} = 0.159 \quad \textbf{Sens.$\downarrow$} = 0.158}
\nodepart{second}
\small The film's center will not \textcolor{red1}{hold} .};
\node[xshift=4.7ex, yshift=-1.8ex, overlay, fill=black, text=white, draw=white, text width=1.35cm, text depth=0.0cm,] at (vapgd2.north west) {VaPGD};
\node[xshift=15.7ex, yshift=1.8ex, overlay, draw=white, text width=5.2cm, text depth=0.0cm,] at (vapgd2.north west) {(a) Model Prediction: Negative};

\node[punkt, below=0cm of vapgd2](intg){
\Textbox{0.3cm}{\small \colorbox{black}{\color{white} IngGrad} \quad \textbf{Comp.} = 0.450 \quad \textbf{Sens.} = 0.192}
\nodepart{second}
\small The film's center will \textcolor{red1}{not} hold .};
\node[xshift=4.7ex, yshift=-1.8ex, overlay, fill=black, text=white, draw=white, text width=1.35cm, text depth=0.0cm,] at (intg.north west) {IngGrad};

\node[punkt, below=0cm of intg](random){
\Textbox{0.3cm}{\small \colorbox{black}{\color{white} Random} \quad \textbf{Comp.} = 0.377 \quad \textbf{Sens.} = 0.252}
\nodepart{second}
\small The film's center \textcolor{red1}{will} not hold .};
\node[xshift=4.7ex, yshift=-1.8ex, overlay, fill=black, text=white, draw=white, text width=1.35cm, text depth=0.0cm,] at (random.north west) {Random};

\node[punx, below=0cm of random](groupa){
\Textbox{0.4cm}{}
};

\node[punkt, below=0cm of groupa](vapgd){
\Textbox{0.3cm}{\small \colorbox{black}{\color{white} VaPGD} \quad \quad \textbf{Comp.$\uparrow$} = 0.184 \quad \textbf{Sens.$\downarrow$} = 4.656}
\nodepart{second}
\small {\textcolor{red1}{Steers}} {\textcolor{red2}{turns}} in a snappy screenplay that curls at the edges ; it 's so clever you want to {\textcolor{red3}{hate}} it.};
\node[xshift=3.9ex, yshift=-1.8ex, overlay, fill=black, text=white, draw=white, text width=1.1cm, text depth=0.0cm,] at (vapgd.north west) {VaPGD};

\node[punkt, below=0cm of vapgd](occlusion){
\Textbox{0.3cm}{\small \colorbox{black}{\color{white} Occlusion} \quad \textbf{Comp.} = 0.552 \quad \textbf{Sens.} = 5.396}
\nodepart{second}
\small {\textcolor{red2}{Steers}} turns in a {\textcolor{red1}{snappy}} screenplay that curls at the edges ; it 's so {\textcolor{red3}{clever}} you want to hate it.};
\node[xshift=5.0ex, yshift=-1.8ex, overlay, fill=black, text=white, draw=white, text width=1.45cm, text depth=0.0cm,] at (occlusion.north west) {Occlusion};
\node[xshift=15.7ex, yshift=1.8ex, overlay, draw=white, text width=5.2cm, text depth=0.0cm,] at (vapgd.north west) {(b) Model Prediction: Positive};

\end{tikzpicture}
\caption{Two examples demonstrating different notions of faithfulness given by Comp. and Sens. A deeper red means the token is identified as more important. Comp. and Sens. scores are also shown.}
\label{bilstmexamplefig1}
\end{figure}

\subsection{Discussion}
\begin{figure*}[h]
\centering
\includegraphics[scale=0.37]{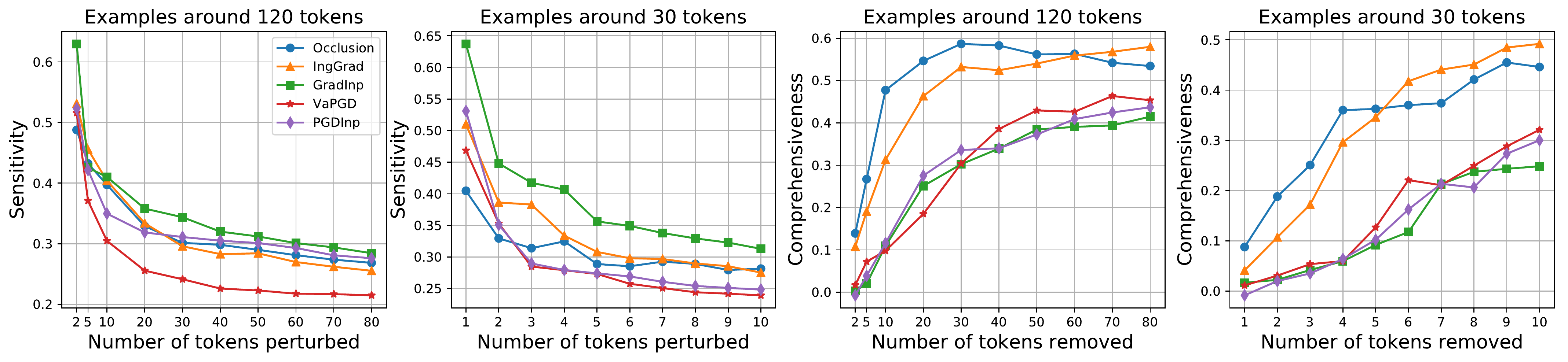}
\caption{Evaluation curves of five interpretation methods. The title of each figure indicates the group of examples based on input lengths. The X-axis is the number of tokens being perturbed or removed for each instance, which varies in 1, 2, $\ldots$, 10 for 30 tokens and 2, 5, 10, 20, $\ldots$, 80 for 120 tokens. The Y-axis is the performance under the criterion. Results imply that IngGrad and VaPGD could be better at identifying a relevant set with more tokens.}
\label{threshold}
\end{figure*}

\paragraph{Performance Curves} To show how the size of the relevant set affects interpretation performance, we plot the comprehensiveness and the sensitivity curves when increasing the number of tokens being removed (perturbed). Consider interpreting BERT on Yelp as an example, we collect two groups of examples from the test set of Yelp based on input lengths, where examples in each group are of 30 $\pm$ 5 and 120 $\pm$ 5 tokens long, and remove (perturb) the top-$k$ most important tokens given by interpretations. Results are shown in Figure \ref{threshold}. 

As shown in the figure, Occlusion is able to discover a smaller set of impactful tokens, under both metrics. However, when the size of the relevant set is increased, the performance of IngGrad under the comprehensiveness metric and the performance of VaPGD under the sensitivity metric gradually surpass Occlusion and other methods. This implies that the two methods are better at identifying a relevant set with more tokens.

\paragraph{Interpolation Analysis}
\begin{figure}[t]
    \centering
    \includegraphics[scale=0.40]{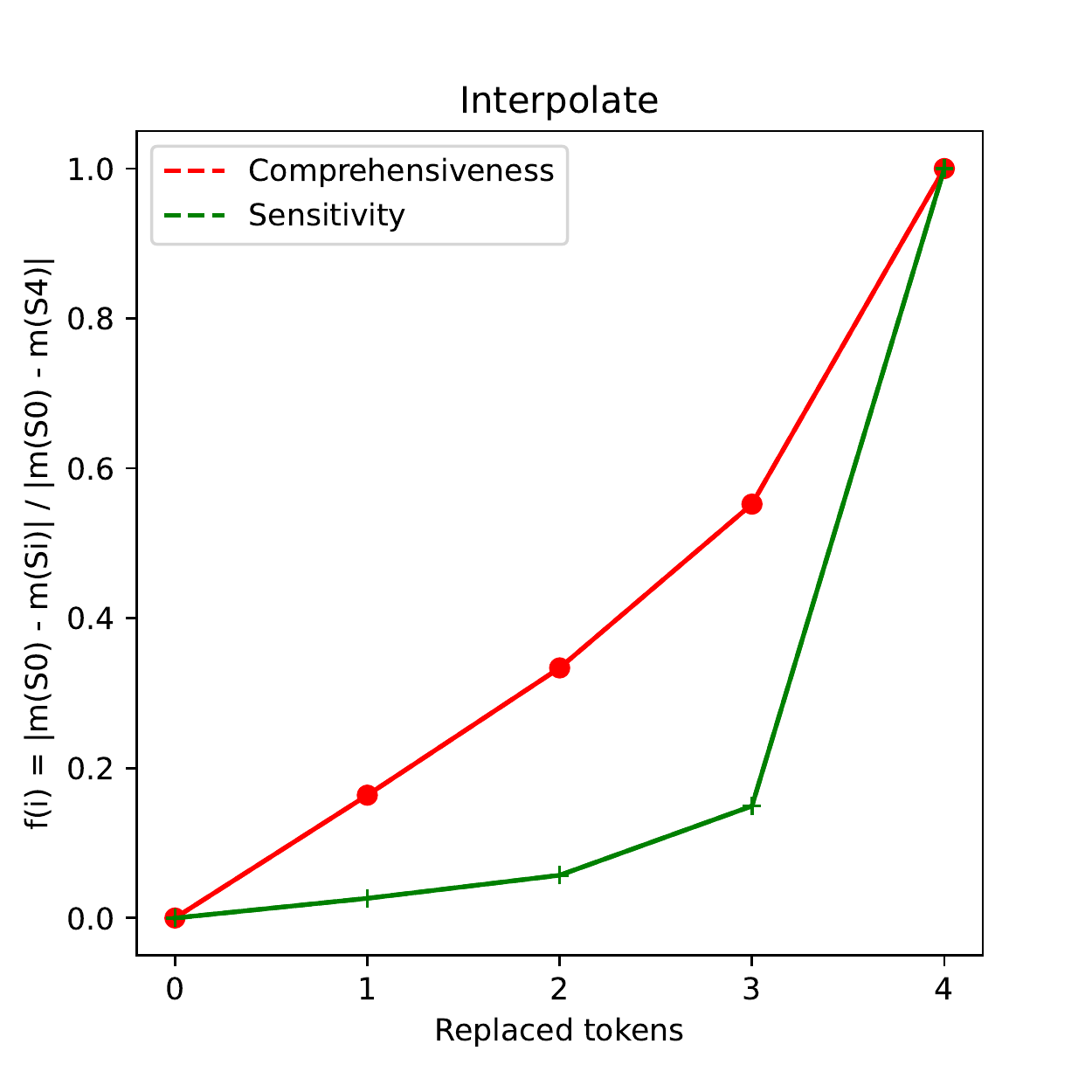}
    \caption{Interpolation between the relevant set and a random set. The relevant set for Comprehensiveness is given by LIME, and the set for Sensitivity is given by VaPGD. We find that Sensitivity better preserves the relative importance for each token in the relevant set.}
    \label{interpolation}
\end{figure}

To check whether the comprehensiveness and sensitivity scores can reflect the relative importance of each token in the relevant set, we conduct an interpolation analysis that gradually replaces each token in the relevant set with a random token outside of the set. 

Specifically, we select 50 examples from SST-2 and test on BERT with relevant sets given by LIME and VaPGD. For each example, we extract a relevant set consists of the top four most important tokens and gradually replace each token, from the least to the most important one, with a random token. We denote the relevant set at each step as $S_0, S_1, ..., S_4$, where $S_0$ is the original relevant set containing the top four tokens and $S_4$ is the set of four random tokens. The performance change at step $i$ is represented by  $f\left(i\right) = \frac{|M\left(S_0\right) - M\left(S_i\right)|}{|M\left(S_0\right) - M\left(S_4\right)|}$, where $M$ is the comprehensiveness or sensitivity score. We expect that a good metric should induce a monotonically increasing function $f$. Further, $f$ should be strictly convex as that indicates the importance of each token is different.

We plot the curve in Figure \ref{interpolation}. Results show that both the comprehensiveness and sensitivity metrics generate a monotonically increasing function, which indicates that they fully consider each token in the relevant set. Also, notice that based on the comprehensiveness metric, the contribution of each token tends to distribute evenly within the relevant set, which contradicts the fact that tokens in the set have different contribution to the prediction, while the importance rank is better preserved based on the sensitivity metric,.

\paragraph{Different Notions of Faithfulness}
Finally, we qualitatively study the notions of faithfulness defined by comprehensiveness (\textit{comp.}) and sensitivity (\textit{sens.}), and discuss two main differences. 

First, \textit{comp.} removes important tokens during evaluations, which could possibly break the interaction between removed tokens and context tokens, and underestimate the importance of context tokens. In Figure \ref{bilstmexamplefig1}(a), the tokens `not' and `hold' together determine the negative sentiment of the sentence. \textit{Sens.} considers both `not' and `hold' as important tokens as one expects. However, \textit{comp.} regards `hold' less important than `will'. 

Second, \textit{sens.} measures token importance by how much model performance would change after `adversarially perturbing' that token. In this sense, both positive and negative pertinent tokens will be deemed important. In contrast, \textit{comp.} only considers positive pertinent ones. In Figure \ref{bilstmexamplefig1}(b), which is predicted as positive, removing the negative verb `hate' would not influence model performance much. However, adversarially perturbing `hate' (e.g. change `hate' to a more negative verb) might change the model prediction from positive to negative. Thus, \textit{sens.} prefers interpretations that identify `hate' as an important token like VaPGD.

The full version of Figure \ref{bilstmexamplefig1}(b) is in Appendix \ref{Appendix5: textbertandlstmexample}. Some contrast examples generated for the stability criterion are presented in Appendix \ref{Appendix 6: appendixDPexample}.

\section{Experiments on Structured Prediction}
\label{DPRes}
Structured prediction tasks are in the center of NLP applications. However, applying interpretation methods and criteria to these tasks are difficult because 1) the required output is a structure instead of a single score. It is hard to define the contribution of each token to a structured output, and 2) compared to text classification tasks, removing parts of the input like what removal-based criteria do, would cause more drastic changes to model predictions as well as the groundtruth. Therefore, existing works often conduct experiments only on binary or multi-class text classification tasks. To remedy these issues, we investigate interpretations for dependency parsing, with an special focus on analyzing how models resolve the PP attachment ambiguity, which avoids interpreting the structured output as a whole. We show that our sensitivity metric is a better metric for dependency parsing as it causes negligible changes to model outputs compared to removal-based metrics.



\subsection{Evaluation Paradigm}
\label{DPparadigm}
Our paradigm focuses on the PP attachment ambiguity, which involves both syntactic and semantics considerations. A dependency parser needs to determine either the preposition in PP attaches to the preceding noun phrase NP (NP-attachment) or the verb phrase VP (VP-attachment) \citep{hindle1993structural}. The basic structure of ambiguity is VP – NP – PP. For example, in the sentence \textit{I saw a cat with a telescope}, a parser uses the semantics of the noun phrase \textit{a telescope} to predict the head of \textit{with}, which is \textit{saw}. If we change \textit{a telescope} to \textit{a tail}, the head of \textit{with} would become the preceding noun \textit{cat}. We will later call nouns in PPs like \textit{telescope} ``disambiguating nouns", as they provide semantic information for a parser to disambiguate PP attachment ambiguity. The main advantage of this paradigm is that disambiguating nouns can be viewed as ``proxy groundtruths'' for faithfulness as parsers must rely on them to make decisions.

\paragraph{Experimental Setup}
\label{DPsetup}
We use DeepBiaffine, a graph-based dependency parser as the target model  \citep{dozat2016deep}. We extract 100 examples that contain the PP attachment ambiguity from the English Penn Treebank converted to Stanford Dependencies 3.5.0 (PTB-SD). We consider the same interpretation methods as before, and they assign an importance score to each token in the sentence to indicate how much it impacts the model prediction on PP attachment arcs. We test the faithfulness of the attributions using 
comprehensiveness and sensitivity.
See Appendix \ref{Appendix1: datastatistics}$\sim$\ref{Appendix3: evaldetails} for details.



\subsection{Results and Discussion}
\label{DPresanddis}
\begin{table}[t]
\renewcommand{\arraystretch}{0.85}
\begin{center}
\small
\begin{tabular}{m{1.8cm}<{\centering}m{2.0cm}<{\centering}m{2.0cm}<{\centering}}
\toprule 
& \multicolumn{2}{c}{\bf{PTB-SD}} \\
\cmidrule{2-3}
\textbf{Method} & \textbf{Comp.} & \textbf{Sens.} \\ 
\toprule 
Random & 0.051 & 10.928 \\
VaGrad &  0.156 & 3.373\\
GradInp &  0.152 & 5.257\\
IngGrad & 0.190 & 4.315\\
DeepLIFT & 0.153 & 5.252\\
Occlusion &  0.194 & 4.671 \\
LIME & \textbf{0.195} & 4.529\\
PGDInp & 0.163 &  4.704 \\
VaPGD & 0.157& \textbf{3.358} \\
Certify & 0.155 & 4.701 \\ 
\bottomrule\hline
\end{tabular}
\end{center}
\caption{Evaluating interpretations for DeepBiaffine under the comprehensiveness and the sensitivity metric on the dependency parsing task.}
\label{DP_results}
\end{table}
Results are shown in Table \ref{DP_results}. Similar to the results on text classification tasks, we find that perturbation-based methods like LIME and Occlusion perform well under the comprehensiveness score, while VaPGD performs the best under sensitivity. PGDInp and Certify are slightly better than GradInp under both the two metrics. 

Qualitatively, we find that according to interpretation methods, important tokens for a PP-attachment decision converge tothe preposition itself, the preceding noun or verb, and the disambiguating noun. This is close to human expectations. An example is shown in Appendix \ref{Appendix5: textbertandlstmexample}.

\begin{table}[t]
\renewcommand{\arraystretch}{0.85}
\begin{center}
\small
\begin{tabular}{m{1.1cm}<{\centering}m{1.1cm}<{\centering}m{1.1cm}<{\centering}m{1.1cm}<{\centering}m{1.1cm}<{\centering}}
\toprule 
& PGD & Occlusion & IngGrad & GradInp\\
\toprule 
Comp. & 0.82 & 0.81 & 0.81 & 0.79\\
Sens. & 0.95 & 0.96 & 0.95 & 0.95\\
\bottomrule\hline
\end{tabular}
\end{center}
\caption{Similarity between the parser outputs before and after applying the evaluation metric. We show that sensitivity changes the global model output less.}
\label{compare2}
\end{table}
\paragraph{Metric Check} Removing even a small piece of inputs breaks the dependency tree. It will be hard to distinguish either the decision process behind the model has changed or the removal of important tokens actually causes the performance drop. Thus, we expect a better metric to have less influence on the tree structure of a sentence. In Table \ref{compare2}, we show that evaluating interpretations with sensitivity leads to smaller changes in the output dependency tree compared to comprehensiveness, suggesting sensitivity a more compatible metric for the dependency parsing task interpretations.

\paragraph{Disambiguating Noun Analysis}
Disambiguating nouns are expected to be identified as important signals by faithful interpretations. We summarize how many times they are actually recognized as the top-k most important words by interpretation methods, where k is the interval varies in 10-20\%, \ldots, 90-100\% of total tokens in an example.

\begin{figure}[t]
\centering
\includegraphics[scale=0.57]{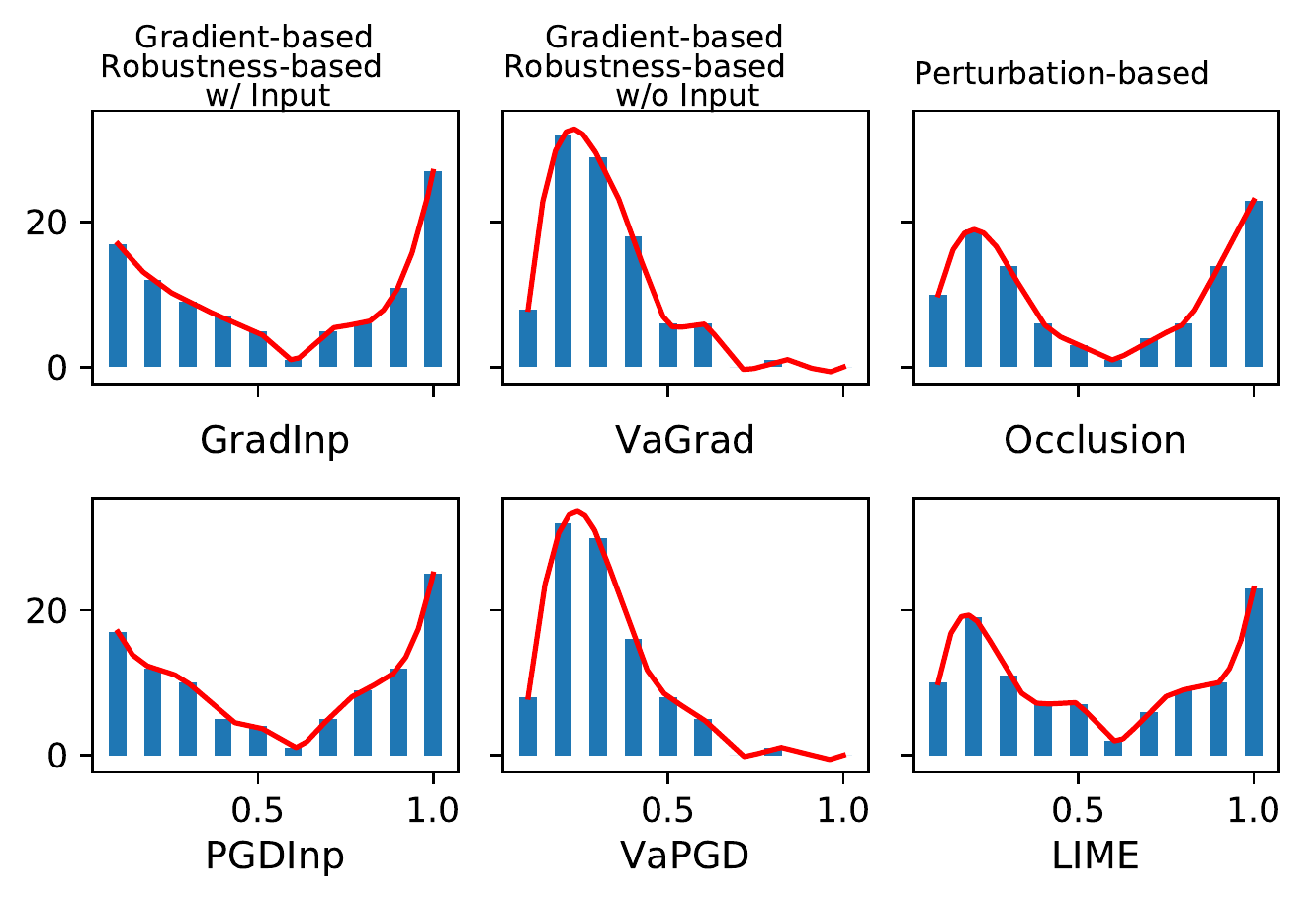}
\caption{Where do interpretations place the disambiguating nouns. The results demonstrate obvious patterns in different categories. The X-axis is the top-k interval. Scales in \{10\%, 20\%, $\ldots$, 100\%\}. The Y-axis is the number of examples that an interpretation ranks the disambiguating noun within each top-k interval.}
\label{np_position}
\end{figure}

Results in Figure \ref{np_position} demonstrate that interpretation methods from the same category have high correlations when extracting disambiguating nouns. For example, VaGrad and VaPGD leveraging gradients only, tend to position disambiguating nouns on the top of their importance lists, which is consistent with human judgments. Likewise, the perturbation-based methods, Occlusion and LIME, also put the disambiguation words to very similar positions.


\section{Related Work}
\noindent{\bf Interpretation methods} Various post-hoc interpretation methods are proposed to explain the behaviors of black-box models. These methods can be roughly categorized into three classes: gradient-based methods \citep{Simonyan13saliency, Li16Visual}, which leverage local gradient information; reference-based methods \citep{Shrikumar17Learning, Sundararajan17Axio}, which consider the model output difference between the original point and a reference point; and perturbation-based methods \citep{Ribeiro2016Why, ZeilerF14, lundberg2017unified}, which query model outputs on perturbed data. In our work, we propose new interpretation methods called robustness-based methods, which adopt techniques in the adversarial robustness domain and bridge the gap between the gradient-based and the reference-based methods.

\noindent{\bf Evaluating interpretation methods} One line of studies explores approaches to evaluate interpretations. Several studies propose measurements for faithfulness. A large proportion of them occlude tokens identified as important by interpretations and measure the confidence change of models \citep{deyoung2020eraser, jain2019attention, zaidan2008modeling, serrano2019is}. Some other works propose to evaluate the faithfulness by checking to what extent they satisfy some desired axioms \citep{AnconaCO018, Sundararajan17Axio, Shrikumar17Learning}. Besides, \citet{alvarez2018robustness, GhorbaniAZ19, kindermans2019reliability, yeh2019fidelity} reveal limitations in interpretation faithfulness through testing the robustness of interpretations. Another group of studies measure the plausibility of interpretations, i.e., whether the explanations conform with human judgments \citep{Doshi2017Towards, Ribeiro2016Why}, or assist humans or student models to predict model behaviors on new data \citep{hase2020evaluating, pruthi2020evaluating}. Note that although there exist many hybrid works that evaluate both the faithfulness and the plausibility of interpretations by combining a suite of diagnostic tests \citep{deyoung2020eraser, atanasova2020diagnostic, liu2020interpretations}, \citet{jacovi2020towards} advocate to explicitly distinguish between the two measurements. In our work, we focus on interpretation faithfulness but consider two new metrics. We apply them to the dependency parsing task. Also, notice that the stability could be regarded as an automatic input consistency tests suggested by \citet{ding2021evaluating}.

\section{Conclusion}
In our work, we enhanced the existed definition of interpretation faithfulness by two other notions and proposed corresponding quantitative metrics: sensitivity and stability,  for each of them. We studied interpretations under the two notions along with the existed one. We found that interpretations have inconsistent performance regarding different criteria. We proposed a new class of interpretations, motivated by the adversarial robustness techniques, which achieves the best performance under the sensitivity and the stability criteria. We further proposed a novel paradigm to evaluate interpretations on the dependency parsing task, which moves beyond text classification in the literature. Our study shed light on understanding the behavior of model interpretations and suggested the community to put more efforts on defining an appropriate evaluation pipeline for interpretation faithfulness.

\section*{Acknowledgment}
We thank anonymous reviewers, UCLA PLUSLab and UCLA-NLP for their helpful feedback. 
This work is supported in part by NSF 1927554, 2008173, 2048280,  CISCO, and a Sloan fellowship.

\section*{Ethical Considerations}
This paper does not contain direct social influences. However, we believe the model analysis and interpretation techniques discussed in this paper are critical for deploying deep learning based models to real-world applications. Following previous work in this direction such as \citet{jacovi2020towards}, we advocate to carefully consider the explanations obtained from interpretation methods as they may not always reflect the true reasoning process behind model predictions. 

Besides the three notions of faithfulness discussed in this paper, there are other important aspects for measuring interpretations that could be applied to evaluate interpretations. Also, We are not claiming that the proposed paradigm are perfect as faithfulness measurements. For example, we recognize that it requires further and detailed analysis on either the model itself or the interpretation methods lead to a low performance on the \emph{stability} metric, although we do try to make sure models behaviors do not change substantially between an input pair. 

Moreover, experiments in this paper are all based on mainstream English corpora. Although our techniques are not language specific, there could be different conclusions given the varying properties of languages. For example, the discussion for dependency parsing could be easily affected by the language one considers.

\bibliography{nlp,ref,newref}
\bibliographystyle{acl_natbib}
\appendix
\clearpage
\input{appendix.tex}

\end{document}

%% file: appendix.tex
\title{Appendix}
\maketitle

\section{Dataset and Model Details}
\label{Appendix1: datastatistics}
\paragraph{Datasets} Statistics of the datasets are presented in Table \ref{DataStatistics}.

\begin{table}[h]
\centering
\begin{tabular}{m{1.38cm}m{3.2cm}<{\centering}m{.75cm}<{\centering}}
  \toprule
  Dataset & Train/Dev/Test & Avg Len \cr
  \toprule
  SST-2 & 67.3k/0.8k/1.8k & 19.2 \cr
  Yelp & 447.9k/112.0k/1.2k & 119.8 \cr
  AGNews & 51.0k/9.0k/3.8k & 35.5 \cr
  PTB-SD & 39.8k/1.7k/2.4k & 23.5 \cr

  \bottomrule\hline
\end{tabular}
\caption{Data Statistics}
\label{DataStatistics}
\end{table}
\paragraph{Models} All models are implemented based on the PyTorch \footnote[1]{https://pytorch.org/} library. All experiments are conducted on NVIDIA GeForce GTX 1080 Ti GPUs.
For BERT, we use the bert-base-uncased model. We fine-tune BERT model on each dataset, using a unified setup: dropout rate 0.1, Adam \citep{kingma2015adam} with an initial learning rate of 1e-4, batch size 128, and no warm-up steps. We set the maximum number of fine-tuning to be 3. The fine-tuned BERT achieves 90.7, 95.4, and 96.9 accuracy on SST-2, Yelp and AGNews, respectively. When explaining BERT predictions, we only consider the contribution of word embeddings to the model output. 

For BiLSTM classifier, we use an one-layer BiLSTM encoder with a linear classifier. The embedding is initialized with the 100-dimensional pre-trained GloVe word embedding. We use Adam with an initial learning rate of 1e-3, batch size 512, hidden size 100 and dropout rate 0.2 for training. We set the maximum number of epochs to be 20 but perform early stopping when the performance on the development set doesn't improve for three epochs. Our BiLSTM classifier receives 84.2, 93.3, 95.9 accuracy on SST-2, Yelp and AGNews, respectively.

For DeepBiaffine, we simplify the original architecture by using a one-layer BiLSTM encoder and a biaffine classifier. The word embedding is also initialized with the pre-trained 100-dimensional GloVe word embedding while the part-of-speech tag embeddings are initialized to all zero. The encoder hidden size is 100. The arc and dependency relation hidden size are both 500. We get an UAS of 95.1 with our model. Note that for DeepBiaffine, each input token is represented by the concatenation of its word embedding and its part-of-speech tag embedding. When applying the interpretation methods and the evaluation metrics, we only modify the word embeddings but keep the part-of-speech tag embeddings unchanged.

\section{Interpretation Methods Details}
\label{Appendix2: interpdetails}
For VaGrad, GradInp, VaPGD, PGDInp, and IngGrad, we use the automatic differentiation mechanism of PyTorch. For LIME, we modify the code from the original implementation of \citet{Ribeiro2016Why} \footnote[1]{https://github.com/marcotcr/lime}. For DeepLIFT, we use the implementation in Captum \footnote[2]{https://github.com/pytorch/captum}. For Certify, we modify the code in auto\_LiRPA \footnote[3]{https://github.com/KaidiXu/auto\_LiRPA}.

For the two reference-based methods IngGrad and DeepLIFT, we use all zero word embeddings as the reference point. To approximate the integral in IngGrad, we sum up 50 points along the linear path from the reference point to the current point. For the perturbation-based methods LIME and Occlusion, we also set the word embedding of a token to an all zero embedding when it is perturbed. 

\paragraph{Hyper-parameter tuning} For all interpretations that require hyper-parameter tuning, including LIME, PGDInp, VaPGD, we randomly select 50 examples from the development set and choose the best hyperparameters based on the performance on these 50 examples. Specifically, the number of perturbed examples around the original point for LIME to fit a linear regression model is selected from \{100, 200, 500, 800\}. For PGDInp and VaPGD, we select the best maximum perturbation norm $\epsilon$ as for BERT and BiLSTM classifier from \{0.1, 0.5, 1.2, 2.2\}. We set the number of iterations as 50, and the step size as $\epsilon / 5$. Note that we might be able to achieve better performance of VaPGD and PGDInp by also tuning the number of iterations and the step size. However, to keep the computational burden comparable with other interpretations, we do not tune these hyperparameters.

\section{Evaluation Criteria Details}
\label{Appendix3: evaldetails}
\paragraph{Sensitivity Details} We use PGD with a binary search for the minimal perturbation magnitude. In practice, we set the number of iterations to be 100 and the step size to be 1.0. Then, we conduct a binary search to estimate the smallest vicinity of the original point which contains an adversarial example that changes the model prediction.
\paragraph{Stability Details}
The synonyms in the stability metrics come from \citep{alzantot2018generating}, where they extract nearest neighbors in the GloVe embeddings space and filter out antonyms with a counter-fitting method. We allow at most four tokens replaced by their synonyms for each input and at most 0.1 change in the output probability of the model prediction for BERT and 0.2 for BiLSTM.

\paragraph{Thresholds} To compute the removal-based metrics and the AUC of sensitivity for text classification tasks, we vary the number of tokens being removed (preserved) or perturbed to be 10\%, 20\%, $\ldots$, 50\% of the total number of tokens in the input. For the dependency parsing task, the corresponding thresholds are 10\%, 20\% and 30\%.

\section{Task Details}
\label{Appendix4: task details}
We evaluate the interpretation methods under both the text classification task and the dependency parsing task. Below, we cover implementation details for each task, respectively, including what is the specific model score interpretation methods explain, and what metrics we use for that task.

\paragraph{Text Classification Task} $s_y\left(e\left(x\right)\right)$ is the probability after the Softmax function corresponding to the original model prediction. We apply all the metrics mentioned in the main paper: removal-based metrics, including comprehensiveness and sufficiency scores, sensitivity score, and stability score. For removal-based metrics, we replace the important tokens with the pad token as a proxy for removing it.

\paragraph{Dependency Parsing Task} $s\left(e\left(x\right)\right)$ is the unlabeled arc log probability between the preposition and its head, i.e., unlabeled arc score after log\_softmax, in the graph-based dependency parser. We discard the sufficiency score as it is unreasonable to remove a large proportion of tokens on a structured prediction task. We also discard the stability metric as there is little consensus on how to attack a structured model.

\section{An Example of Interpreting BERT and BiLSTM on the Text Classification Task}
\label{Appendix5: textbertandlstmexample}
We showcase an example for interpreting BERT and BiLSTM in Figure \ref{appendBertTextExample} and \ref{bilstmexample}. The example comes from the test set of SST-2. A deeper red color means the token is identified as more important to the model output by an interpretation while a deeper blue color stands for a less important token. Both the BiLSTM classifier and BERT classifier assign a positive label to this instance. Qualitatively, given an input, we observe that the most relevant or irrelevant sets of words identified by different interpretations are highly overlapped for BiLSTM, although the exact order of importance scores might be different. Whereas for BERT, different interpretations usually give different important tokens.

\colorlet{red1}{red}
\colorlet{red2}{red!50}
\colorlet{red3}{red!25}
\colorlet{blue1}{blue}
\colorlet{blue2}{blue!50}
\colorlet{blue3}{blue!25}
\makeatletter
\pgfdeclaregenericanchor{top base}{%
  \csname pgf@anchor@#1@north\endcsname
  \pgf@anchor@generic@top@base@main
}
\pgfdeclaregenericanchor{top base west}{%
  \csname pgf@anchor@#1@north west\endcsname
  \pgf@anchor@generic@top@base@main
}
\pgfdeclaregenericanchor{top base east}{%
  \csname pgf@anchor@#1@north east\endcsname
  \pgf@anchor@generic@top@base@main
}
\def\pgf@anchor@generic@top@base@main{%
  {%
    \pgfmathsetlength\pgf@ya{\pgfkeysvalueof{/pgf/outer ysep}}%
    \advance\pgf@y-\pgf@ya
    \pgfmathsetlength\pgf@ya{\pgfkeysvalueof{/pgf/inner ysep}}%
    \advance\pgf@y-\pgf@ya
    \pgf@ya=0pt
    \pgfutil@loop
    \ifdim\pgf@y>\baselineskip
      \advance\pgf@y-\baselineskip
      \advance\pgf@ya\baselineskip
    \pgfutil@repeat
    \global\pgf@y=\pgf@ya
  }%
}
\makeatother

\begin{figure}[!t]

\tikzset{
    >=stealth',
    punkt/.style={
          very thick,
          rectangle split,
          rectangle split parts=2,
          inner ysep=1.2mm,
          inner xsep=1.5mm,
          rounded corners,
          draw=black, thick,
          text width=7.2cm},
    punx/.style={
          rectangle,
          rounded corners,
          text width=0.55cm,
          text depth=0.15cm,
          draw=white},
  header/.style = {%
    inner ysep = +1.5em,
    append after command = {
      \pgfextra{\let\TikZlastnode\tikzlastnode}
      node [header node] (header-\TikZlastnode) at (\TikZlastnode.north) {#1}
      node [span = (\TikZlastnode)(header-\TikZlastnode)]
        at (fit bounding box) (h-\TikZlastnode) {}
    }
  },
}

\colorlet{red1}{red}
\colorlet{red2}{red!50}
\colorlet{red3}{red!25}
\colorlet{blue1}{blue}
\colorlet{blue2}{blue!50}
\colorlet{blue3}{blue!25}
\begin{tikzpicture}[scale=1.8]

\node[punkt](pgd){
\Textbox{0.3cm}{\small \colorbox{black}{\color{white} PGDInp} \quad \textbf{Comp.} = 0.776 \quad \textbf{Sens.} = 0.349}
\nodepart{second}
\small \textcolor{blue3}{Steers} turns in \textcolor{blue2}{a} snappy \textcolor{red2}{screenplay} that curls at the edges ; it \textcolor{blue1}{'s} so \textcolor{red1}{clever} you want to \textcolor{red3}{hate} it.};
\node[xshift=4.5ex, yshift=-1.8ex, overlay, fill=black, text=white, draw=white] at (current bounding box.north west) { PGDInp};

\node[punkt, below=0cm of pgd](vapgd){
\Textbox{0.3cm}{\small \colorbox{black}{\color{white} VaPGD} \quad \textbf{Comp.} = 0.759 \quad \textbf{Sens.} = 0.339}
\nodepart{second}
\small \textcolor{blue2}{Steers} turns in \textcolor{blue3}{a} snappy \textcolor{red3}{screenplay} that curls \textcolor{blue1}{at} the edges ; it 's so \textcolor{red1}{clever} you want to \textcolor{red2}{hate} it.};
\node[xshift=4.0ex, yshift=-1.8ex, overlay, fill=black, text=white, draw=white] at (vapgd.north west) { VaPGD};

\node[punkt, below=0cm of vapgd](occlusion){
\Textbox{0.3cm}{\small \colorbox{black}{\color{white} Occlusion} \quad \small \textbf{Comp.} = 0.962 \quad \textbf{Sens.} = 0.376}
\nodepart{second}
\small Steers turns in a snappy \textcolor{blue3}{screenplay} that curls at the edges ; it \textcolor{red3}{'s} \textcolor{red2}{so} \textcolor{red1}{clever} you \textcolor{blue2}{want} to \textcolor{blue1}{hate} it.};
\node[xshift=5ex, yshift=-1.8ex, overlay, fill=black, text=white, draw=white] at (occlusion.north west) {Occlusion};


\node[punkt, below=0cm of occlusion](intg){
\Textbox{0.3cm}{\small \colorbox{black}{\color{white} IngGrad} \quad \textbf{Comp.} = 0.930 \quad \textbf{Sens.} = 0.383}
\nodepart{second}
\small Steers turns in a snappy \textcolor{blue1}{screenplay} \textcolor{red3}{that} curls at the edges \textcolor{red2}{;} it 's so \textcolor{red1}{clever} you \textcolor{blue3}{want} to \textcolor{blue2}{hate} it.};
\node[xshift=4.7ex, yshift=-1.8ex, overlay, fill=black, text=white, draw=white] at (intg.north west) { IngGrad};

\node[punkt, below=0cm of intg](gradinp){
\Textbox{0.3cm}{\small \colorbox{black}{\color{white} GradInp} \textbf{Comp.} = 0.907 \quad \textbf{Sens.} = 0.352}
\nodepart{second}
\small Steers turns in \textcolor{blue3}{a} snappy \textcolor{red2}{screenplay} that curls at the edges \textcolor{red3}{;} it \textcolor{blue2}{'s} so \textcolor{red1}{clever} you \textcolor{blue1}{want} to hate it.};
\node[xshift=4.6ex, yshift=-1.8ex, overlay, fill=black, text=white, draw=white] at (gradinp.north west) {GradInp};

\end{tikzpicture}
\caption{An example of interpreting BERT with five interpretation methods. A deeper red color means the token is identified as more important while a deeper blue color stands for a less important token. Performance under Comp. and Sens. scores are shown.}
\label{appendBertTextExample}
\end{figure}

\section{An Example of Interpreting the Dependency Parser}
\label{Appendix 6: appendixDPexample}

An example of interpreting the PP attachment decision of a DeepBiaffine model. A deeper red color means the token is identified as more important for the model to predict the PP attachment arc.

\begin{figure}[!t]

\tikzset{
    >=stealth',
    punkt/.style={
           very thick,
           rectangle split,
           rectangle split parts=2,
           inner ysep=1.2mm,
           inner xsep=1.5mm,
           rounded corners,
           draw=black, thick,
           text width=7.2cm},
    punx/.style={
           rectangle,
           rounded corners,
           text width=0.55cm,
           text depth=0.15cm,
           draw=white},
  header/.style = {%
    inner ysep = +1.5em,
    append after command = {
      \pgfextra{\let\TikZlastnode\tikzlastnode}
      node [header node] (header-\TikZlastnode) at (\TikZlastnode.north) {#1}
      node [span = (\TikZlastnode)(header-\TikZlastnode)]
        at (fit bounding box) (h-\TikZlastnode) {}
    }
  },
}

\begin{tikzpicture}[scale=1.0]

\node[punkt](pgd){
\Textbox{0.3cm}{\small \colorbox{black}{\color{white} PGDInp} \quad \textbf{Comp.} = 0.550 \quad \textbf{Sens.} = 5.203}
\nodepart{second}
\small {\textcolor{red2}{Steers}} turns in a {\textcolor{red1}{snappy}} {\textcolor{blue2}{screenplay}} that curls at the edges ; {\textcolor{blue3}{it}} 's so {\textcolor{red3}{clever}} you want to {\textcolor{blue1}{hate}} it.};
\node[xshift=4.5ex, yshift=-1.8ex, overlay, fill=black, text=white, draw=white] at (current bounding box.north west) { PGDInp};

\node[punkt, below=0cm of pgd](vapgd){
\Textbox{0.3cm}{\small \colorbox{black}{\color{white} VaPGD} \quad \quad \textbf{Comp.} = 0.184 \quad \textbf{Sens.} = 4.656}
\nodepart{second}
\small {\textcolor{red1}{Steers}} {\textcolor{red2}{turns}} in a snappy screenplay that curls {\textcolor{blue3}{at}} {\textcolor{blue2}{the}} edges ; it {\textcolor{blue1}{'s}} so clever you want to {\textcolor{red3}{hate}} it.};
\node[xshift=4.0ex, yshift=-1.8ex, overlay, fill=black, text=white, draw=white, text width=1.1cm, text depth=0.0cm,] at (vapgd.north west) {VaPGD};

\node[punkt, below=0cm of vapgd](occlusion){
\Textbox{0.3cm}{\small \colorbox{black}{\color{white} Occlusion} \quad \textbf{Comp.} = 0.552 \quad \textbf{Sens.} = 5.396}
\nodepart{second}
\small {\textcolor{red2}{Steers}} turns in a {\textcolor{red1}{snappy}} screenplay that curls at the edges ; {\textcolor{blue2}{it}} 's so {\textcolor{red3}{clever}} you {\textcolor{blue3}{want}} to {\textcolor{blue1}{hate}} it.};
\node[xshift=5.0ex, yshift=-1.8ex, overlay, fill=black, text=white, draw=white, text width=1.45cm, text depth=0.0cm,] at (occlusion.north west) {Occlusion};

\node[punkt, below=0cm of occlusion](intg){
\Textbox{0.3cm}{\small \colorbox{black}{\color{white} IngGrad} \quad \textbf{Comp.} = 0.609 \quad \textbf{Sens.} = 5.310}
\nodepart{second}
\small {\textcolor{red2}{Steers}} turns in a {\textcolor{red1}{snappy}} {\textcolor{blue3}{screenplay}} that curls at the edges ; it 's so {\textcolor{red3}{clever}} you {\textcolor{blue2}{want}} to {\textcolor{blue1}{hate}} it.};
\node[xshift=4.7ex, yshift=-1.8ex, overlay, fill=black, text=white, draw=white, text width=1.35cm, text depth=0.0cm,] at (intg.north west) {IngGrad};

\node[punkt, below=0cm of intg](gradinp){
\Textbox{0.3cm}{\small \colorbox{black}{\color{white} GradInp} \quad \textbf{Comp.} = 0.546 \quad \textbf{Sens.} = 5.304}
\nodepart{second}
\small {\textcolor{red3}{Steers}} turns in a {\textcolor{red1}{snappy}} screenplay that curls at the edges ; {\textcolor{blue2}{it}} 's so {\textcolor{red2}{clever}} you {\textcolor{blue3}{want}} to {\textcolor{blue1}{hate}} it.};
\node[xshift=4.6ex, yshift=-1.8ex, overlay, fill=black, text=white, draw=white, text width=1.35cm, text depth=0.0cm,] at (gradinp.north west) {GradInp};

\end{tikzpicture}
\caption{An example of interpreting BiLSTM using five interpretation methods.}
\label{bilstmexample}
\end{figure}

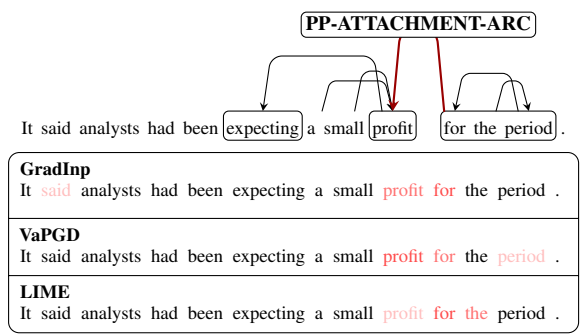
\begin{figure}[h]

    \centering
\begin{dependency}[hide label, theme = default]
\tikzstyle{word}=[text=black, anchor=north east, inner sep=0.2ex,
font=\scriptsize]
  \begin{deptext}[column sep=0.1em]
  |[word]|It \& |[word]| said \& |[word]| analysts \& |[word]| had \& |[word]| been \& |[word]| expecting \& |[word]| a \& |[word]| small \& |[word]| profit \&[.3cm] |[word]| for \& |[word]| the \&|[word]|  period \& |[word]| .\\
  \end{deptext}
  \depedge[edge height=0.4cm]{7}{9}{det}
  \depedge[edge height=0.53cm]{8}{9}{amod}
  \depedge[edge height=0.73cm]{9}{6}{dobj}
  \depedge[edge height=1.13cm, edge above, show label, edge style={red!60!black, thick},
label style={font=\bfseries,text=black}]{10}{9}{PP-ATTACHMENT-ARC}
  \depedge[edge height=0.4cm]{11}{12}{det}
  \depedge[edge height=0.5cm]{12}{10}{pobj}
  \wordgroup{1}{6}{6}{verb}
  \wordgroup{1}{9}{9}{noun}
  \wordgroup{1}{10}{12}{pp}
\end{dependency}

\tikzset{
    >=stealth',
    punkt/.style={
          rectangle split,
          rectangle split parts=3,
          rounded corners,
          draw=black,
          inner ysep=1.2mm,
          inner xsep=1.5mm,
          text width=7.2cm,
          text depth=0.45cm},
    punx/.style={
          rectangle,
          rounded corners,
          text width=0.55cm,
          text depth=0.15cm,
          draw=white},
  header/.style = {%
    inner ysep = +1.5em,
    append after command = {
      \pgfextra{\let\TikZlastnode\tikzlastnode}
      node [header node] (header-\TikZlastnode) at (\TikZlastnode.north) {#1}
      node [span = (\TikZlastnode)(header-\TikZlastnode)]
        at (fit bounding box) (h-\TikZlastnode) {}
    }
  },
}
\begin{tikzpicture}
\node[punkt](pgd){
\scriptsize \textbf{GradInp} \\
\scriptsize It {\color{red!25}{said}} analysts had been expecting a small {\color{red!55}{profit}} {\color{red!70}{for}} the period .\\
\nodepart{second}
\scriptsize \textbf{VaPGD} \\
\scriptsize It said analysts had been expecting a small {\color{red!70}{profit}} {\color{red!55}{for}} the {\color{red!25}{period}} .\\
\nodepart{third}
\scriptsize \textbf{LIME} \\
\scriptsize It said analysts had been expecting a small {\color{red!25}{profit}} {\color{red!70}{for}} {\color{red!55}{the}}  period .\\
};
\end{tikzpicture}
    \caption{An example of interpreting the PP attachment arc in the dependency parsing task. A deeper red color means the token is identified as more important for the model to predict the PP attachment arc.}
    \label{DPExample}
\end{figure}

\section{Examples for the Stability Criterion}
\label{Appendix7: stabexm}
\subsection{SST-2 Examples}
Table \ref{stabilityexamples} shows some contrast examples constructed for the stability criterion on SST-2.
\begin{table}[t]
\centering

\begin{tabular}{p{1.3cm}p{5.4cm}}
\toprule
\multicolumn{2}{c}{\bf{VaPGD, BERT on SST-2}} \cr
\midrule
\multicolumn{2}{c}{\bf Rank correlation = 0.346} \cr
\multicolumn{2}{c}{\bf Model change = 0.00} \cr
\textbf{Original} & This is a film well worth seeing , talking and singing heads and all .\cr
\textbf{Contrast} & This is a \underline{films} well worth \underline{staring} , talking and singing heads and \underline{entirety} .\cr

\bottomrule

\toprule
\end{tabular}

\begin{tabular}{p{1.3cm}p{5.4cm}}
\toprule
\multicolumn{2}{c}{\bf{IngGrad, BERT on SST-2}} \cr
\midrule
\multicolumn{2}{c}{\bf Rank correlation = 0.645} \cr
\multicolumn{2}{c}{\bf Model change = 0.15} \cr
\textbf{Original} & Ray Liotta and Jason Patric do some of their best work in their underwritten roles , but do n't be fooled : Nobody deserves any prizes here .\cr
\textbf{Contrast} & Ray Liotta and Jason Patric do \underline{certain} of their best \underline{collaborate} in their underwritten roles , but do n't be fooled : Nobody deserves any \underline{awards} here .\cr

\bottomrule

\toprule
\end{tabular}

\begin{tabular}{p{1.3cm}p{5.4cm}}
\toprule
\multicolumn{2}{c}{\bf{LIME, BiLSTM on SST-2}} \cr
\midrule
\multicolumn{2}{c}{\bf Rank correlation = 0.425} \cr
\multicolumn{2}{c}{\bf Model change = 0.05} \cr
\textbf{Original} & Nearly surreal , dabbling in French , this is no simple movie , and you 'll be taking a risk if you choose to see it .\cr
\textbf{Contrast} & \underline{Almost} surreal , dabbling in French , this is no simple \underline{cinematography} , and you 'll be taking a risk if you choose to \underline{seeing} it .\cr

\bottomrule

\toprule
\end{tabular}


\caption{Generated contrast examples for evaluating the stability criterion on SST-2. Modified words are underlined. Spearman's rank correlation between a pair of examples and the performance difference of a model on the pair of examples are shown above each pair.}
\label{stabilityexamples}
\end{table}

\subsection{AGNews Examples}
Table \ref{stabilityagnewsexamples} shows some contrast examples constructed for the stability criterion on AGNews
\begin{table}[t]
\centering

\begin{tabular}{p{1.3cm}p{5.4cm}}
\toprule
\multicolumn{2}{c}{\bf{Erasure, BERT on AGNew}} \cr
\midrule
\multicolumn{2}{c}{\bf Rank correlation = 0.689} \cr
\multicolumn{2}{c}{\bf Model change = 0.08} \cr
\textbf{Original} & Supporters and rivals warn of possible fraud ; government says chavez 's defeat could produce turmoil in world oil market .\cr
\textbf{Contrast} & Supporters and rivals warn of possible fraud ; government says chavez 's defeat could produce \underline{disorder} in \underline{planet} oil \underline{trade} .\cr

\bottomrule

\toprule
\end{tabular}

\begin{tabular}{p{1.3cm}p{5.4cm}}
\toprule
\multicolumn{2}{c}{\bf{DeepLIFT, BERT on AGNews}} \cr
\midrule
\multicolumn{2}{c}{\bf Rank correlation = 0.317} \cr
\multicolumn{2}{c}{\bf Model change = 0.00} \cr
\textbf{Original} & Mills corp. agreed to purchase a qqq percent interest in nine malls owned by general motors asset management corp. for just over qqq billion , creating a new joint venture between the groups .
\cr
\textbf{Contrast} & Mills corp. \underline{agree} to purchase a qqq percent interest in nine malls owned by \underline{comprehensive} motors asset management corp. for just over qqq \underline{trillion} , creating a new joint venture between the groups .
\cr

\bottomrule

\toprule
\end{tabular}

\begin{tabular}{p{1.3cm}p{5.4cm}}
\toprule
\multicolumn{2}{c}{\bf{VaGrad, BERT on AGNews}} \cr
\midrule
\multicolumn{2}{c}{\bf Rank correlation = 0.970} \cr
\multicolumn{2}{c}{\bf Model change = 0.12} \cr
\textbf{Original} & London ( reuters ) - oil prices surged to a new high of qqq a barrel on wednesday after a new threat by rebel militia against iraqi oil facilities and as the united states said inflation had stayed in check despite rising energy costs .\cr
\textbf{Contrast} & london ( reuters ) - oil prices surged to a new high of qqq a \underline{canon} on wednesday after a new \underline{menace} by rebel militia against iraqi oil facilities and as the united states said inflation had stayed in check despite rising energy costs .\cr

\bottomrule

\toprule
\end{tabular}


\caption{Generated contrast examples for evaluating the stability criterion on AGNews.}
\label{stabilityagnewsexamples}
\end{table}

\subsection{Yelp Examples}
Table \ref{stabilityyelpexamples} shows some contrast examples constructed for the stability criterion on Yelp.
\begin{table}[t]
\centering

\begin{tabular}{p{1.4cm}p{5.4cm}}
\toprule
\multicolumn{2}{c}{\bf{PGD, BiLSTM on Yelp}} \cr
\midrule
\multicolumn{2}{c}{\bf Rank correlation = 0.530} \cr
\multicolumn{2}{c}{\bf Model change = 0.00} \cr
\textbf{Original} & Love this beer distributor. They always have what I'm looking for. The workers are extremely nice and always willing to help. Best one I've seen by far.\cr
\textbf{Contrast} & Love this beer distributor. They \underline{repeatedly} have what I'm \underline{seeking} for. The workers are extremely nice and always \underline{loan} to help. Best one I've seen by far.\cr

\bottomrule

\toprule
\end{tabular}

\begin{tabular}{p{1.4cm}p{5.4cm}}
\toprule
\multicolumn{2}{c}{\bf{Certify, BiLSTM on Yelp}} \cr
\midrule
\multicolumn{2}{c}{\bf Rank correlation = 0.633} \cr
\multicolumn{2}{c}{\bf Model change = 0.01} \cr
\textbf{Original} & Last summer I had an appointment to get new tires and had to wait a super long time. I also went in this week for them to fix a minor problem with a tire they put on. They "fixed" it for free, and the very next morning I had the same issue. I called to complain, and the "manager" didn't even apologize!!! So frustrated. Never going back.  They seem overpriced, too. \cr
\textbf{Contrast} & Last summer I \underline{took} an \underline{appoints} to get new tires and had to wait a super long time. I also went in this week for them to fix a minor problem with a tire they put on. They "fixed" it for free, and the very \underline{impending} morning I had the same issue. I called to complain, and the "manager" didn't even apologize!!! So frustrated. Never going back.  They seem overpriced, too. \cr

\bottomrule

\toprule
\end{tabular}


\caption{Generated contrast examples for evaluating the stability criterion on Yelp.}
\label{stabilityyelpexamples}
\end{table}

\section{Case Study on Gradient Saturation}
\label{Appendix8 Case study Saturation}
We qualitatively study some cases where PGDInp does well under the removal-based criterion while GradInp does not. In Figure \ref{saturationexamples}, we show an example from explaining BERT on the SST-2 dataset, with the importance scores given by PGDInp, VaPGD, GradInp, VaGrad and the comprehensiveness score. For PGDInp and GradInp, we show the exponential of importance scores.

As shown in Figure \ref{saturationexamples}, the importance score for each token given by GradInp is close to zero. VaGrad also gives near zero importance scores. At the same time, PGDInp and VaPGD have distinguishable and meaningful importance scores.

Based on the above observations, we suspect that the reason why PGD-based methods could avoid the failure of gradient-based methods is that they do not suffer from the gradient saturation issue. Gradient saturation refers to the cases where gradients are close to zero at some specific inputs and provide no information about the importance of different features of those inputs. Note that PGD-based methods consider not only a single input, but search on the vicinity of that input where the neighbors have none-zero gradients.

However, notice that VaGrad works better than GradInp. We suspect that is because although all elements in the gradient vector are close to zero, the **L-2 norm** of it is still distinguishable. However, GradInp takes the **dot-product** between embeddings and their gradients as the importance score. It is likely that negative and positive dimensions are neutralized, making the importance scores undistinguishable, and thus the behavior of GradInp corrupted. This hypothesis needs further explorations and demonstrations.

\begin{table}[t]
\small
\begin{center}
\begin{tabular}{l}
Example: A very funny movie .
\end{tabular}
\begin{tabular}{m{0.8cm}<{\centering}m{0.8cm}<{\centering}|m{0.7cm}<{\centering}m{0.7cm}<{\centering}m{0.7cm}<{\centering}m{0.7cm}<{\centering}m{0.7cm}<{\centering}}

\toprule
 &  & \multicolumn{5}{c}{\bf Importance Scores} \cr
\cmidrule{3-7}
\bf Method & \bf Comp. & A & very & funny & movie &.\cr
\midrule
PGDInp & 0.90 & 0.996 & 1.009 & 1.055 & 0.999 & 0.994 \cr
GradInp & 0.33  & 1.000 & 1.000 & 1.000 & 1.000 & 1.000 \cr
VaPGD & 0.67 & 0.072 & 0.124 & 0.399 & 0.199 & 0.079  \cr
VaGrad & 0.54 & 0.000 & 0.001 & 0.001 & 0.001 & 0.000  \cr
\bottomrule
\end{tabular}
\end{center}
\caption{An example showing the gradient saturation issue. We show the importance score for each word given by the four interpretations and the corresponding comprehensiveness score. We find that while gradient-based methods suffer from the saturation issue, PGDInp and VaPGD could avoid the limitation.}
\label{saturationexamples}
\end{table}

\section{Statistical Testing on the Performance of Robustness-based Methods}
We use Student's t test to exam the superior performance of the best robustness-based methods against previous methods. The double checkmark represents a confidence level of 95\% and a single checkmark for a confidence level of 90\%.
\label{statisticaltesting}
\begin{table}[t]
\small
\renewcommand{\arraystretch}{0.85}
\begin{center}

\begin{tabular}{m{0.8cm}<{\centering}m{0.6cm}<{\centering}m{0.65cm}<{\centering}m{0.5cm}<{\centering}m{0.4cm}<{\centering}m{0.65cm}<{\centering}m{0.65cm}<{\centering}}
\toprule 
BERT Yelp & VaGrad & GradInp & Occlu. & LIME & IngGrad & DeepLI.\\
\toprule 
Sensitivity & \checkmark & \checkmark\checkmark & \checkmark & \checkmark\checkmark & \checkmark\checkmark & \checkmark\checkmark\\
Stability & \checkmark & \checkmark\checkmark & \checkmark\checkmark & \checkmark\checkmark & \checkmark & \checkmark\checkmark\\
\bottomrule\hline
\end{tabular}
\begin{tabular}{m{0.8cm}<{\centering}m{0.6cm}<{\centering}m{0.65cm}<{\centering}m{0.5cm}<{\centering}m{0.4cm}<{\centering}m{0.65cm}<{\centering}m{0.65cm}<{\centering}}
BERT SST-2 & VaGrad & GradInp & Occlu.& LIME & IngGrad & DeepLI.\\
\toprule 
Sensitivity & x & \checkmark\checkmark & \checkmark\checkmark & \checkmark\checkmark & \checkmark\checkmark & \checkmark\checkmark\\
Stability & x & \checkmark\checkmark & \checkmark\checkmark & \checkmark\checkmark & \checkmark & \checkmark\checkmark\\
\bottomrule\hline
\end{tabular}
\begin{tabular}{m{0.8cm}<{\centering}m{0.6cm}<{\centering}m{0.65cm}<{\centering}m{0.5cm}<{\centering}m{0.4cm}<{\centering}m{0.65cm}<{\centering}m{0.65cm}<{\centering}}
BERT agnews & VaGrad & GradInp & Occlu. & LIME & IngGrad & DeepLI.\\
\toprule 
Sensitivity & \checkmark & \checkmark\checkmark & \checkmark\checkmark & \checkmark\checkmark & \checkmark\checkmark & \checkmark\checkmark\\
Stability & \checkmark & \checkmark\checkmark & \checkmark\checkmark & \checkmark\checkmark & \checkmark\checkmark & \checkmark\checkmark\\
\bottomrule\hline
\end{tabular}
\end{center}
\caption{Similarity between the parser outputs before and after applying the evaluation metric. We show that sensitivity changes the global model output less. Occlu. and DeepLI. represents Occlusion and DeepLIFT, respectively}
\label{stats}
\end{table}

\begin{table}[t]
\renewcommand{\arraystretch}{0.85}
\begin{center}
\small
\begin{tabular}{m{0.8cm}<{\centering}m{0.6cm}<{\centering}m{0.65cm}<{\centering}m{0.5cm}<{\centering}m{0.4cm}<{\centering}m{0.65cm}<{\centering}m{0.65cm}<{\centering}}
\toprule 
LSTM Yelp & VaGrad & GradInp & Occlu. & LIME & IngGrad & DeepLI.\\
\toprule 
Sensitivity & \checkmark\checkmark & \checkmark\checkmark & \checkmark\checkmark & \checkmark\checkmark & \checkmark\checkmark & \checkmark\checkmark\\
Stability & x & \checkmark\checkmark & \checkmark\checkmark & \checkmark\checkmark & \checkmark\checkmark & \checkmark\checkmark\\
\bottomrule\hline
\end{tabular}
\begin{tabular}{m{0.8cm}<{\centering}m{0.6cm}<{\centering}m{0.65cm}<{\centering}m{0.5cm}<{\centering}m{0.4cm}<{\centering}m{0.65cm}<{\centering}m{0.65cm}<{\centering}}
LSTM SST-2 & VaGrad & GradInp & Occlu. & LIME & IngGrad & DeepLI.\\
\toprule 
Sensitivity & x & \checkmark\checkmark & \checkmark\checkmark & \checkmark\checkmark & \checkmark\checkmark & \checkmark\checkmark\\
Stability & \checkmark & \checkmark\checkmark & \checkmark\checkmark & \checkmark\checkmark & \checkmark\checkmark & \checkmark\checkmark\\
\bottomrule\hline
\end{tabular}
\begin{tabular}{m{0.8cm}<{\centering}m{0.6cm}<{\centering}m{0.65cm}<{\centering}m{0.5cm}<{\centering}m{0.4cm}<{\centering}m{0.65cm}<{\centering}m{0.65cm}<{\centering}}
LSTM agnews & VaGrad & GradInp & Occlu. & LIME & IngGrad & DeepLI.\\
\toprule 
Sensitivity &\checkmark\checkmark & \checkmark\checkmark & \checkmark\checkmark & \checkmark\checkmark & \checkmark\checkmark & \checkmark\checkmark\\
Stability & x & \checkmark\checkmark & \checkmark\checkmark & \checkmark\checkmark & x & \checkmark\checkmark\\
\bottomrule\hline
\end{tabular}
\end{center}
\caption{Similarity between the parser outputs before and after applying the evaluation metric. We show that sensitivity changes the global model output less. Occlu. and DeepLI. represents Occlusion and DeepLIFT, respectively}
\label{stats}
\end{table}